# MCVI-SANet: A lightweight semi-supervised model for LAI and SPAD estimation of winter wheat under vegetation index saturation

Zhiheng Zhang[a], Jiajun Yang[a], Hong Sun[b], Dong Wang[c], Honghua Jiang[a,*], Yaru Chen[d], Tangyuan Ning[a,*]

[a] *College of Information Science and Engineering, Shandong Agricultural University, Taian 271018, China*

[b] *Key Lab of Smart Agriculture Systems, Ministry of Education, China Agricultural University, Beijing 100083, China*

[c] *College of Agriculture, Northwest Agriculture and Forestry University, Yangling 712100, China*

[d] *Centre for Vision Speech and Signal Processing (CVSSP), University of Surrey, Surrey GU2 7XH, United Kingdom*

**Abstract** Vegetation index (VI) saturation during the dense canopy stage and limited ground-truth annotations of winter wheat constrain accurate estimation of LAI and SPAD. Existing VI-based and texture-driven machine learning methods exhibit limited feature expressiveness. In addition, deep learning baselines suffer from domain gaps and high data demands, which restrict their generalization. Therefore, this study proposes the Multi-Channel Vegetation Indices Saturation Aware Net (MCVI-SANet), a lightweight semi-supervised vision model. The model incorporates a newly designed Vegetation Index Saturation-Aware Block (VI-SABlock) for adaptive channel-spatial feature enhancement. It also integrates a VICReg-based semi-supervised strategy to further improve generalization. Datasets were partitioned using a vegetation height-informed strategy to maintain representativeness across growth stages. Experiments over 10 repeated runs demonstrate that MCVI-SANet achieves state-of-the-art accuracy. The model attains an average $R^2$ of 0.8123 and RMSE of 0.4796 for LAI, and an average $R^2$ of 0.6846 and RMSE of 2.4222 for SPAD. This performance surpasses the best-performing baselines, with improvements of 8.95% in average LAI $R^2$ and 8.17% in average SPAD $R^2$. Moreover, MCVI-SANet maintains high inference speed with only 0.10M parameters. Overall, the integration of semi-supervised learning with agronomic priors provides a promising approach for enhancing remote sensing-based precision agriculture.

---

* Corresponding author. E-mail address: j_honghua@sdau.edu.cn (H. Jiang), ningty@sdau.edu.cn (T. Ning).

# 1. Introduction

Wheat is a staple crop that plays a crucial role in ensuring global food security. As the global population continues to rise and the impacts of climate change become more pronounced, enhancing wheat yield and resource-use efficiency is increasingly vital (Lidwell-Durnin & Lapthorn, 2020). In this context, the leaf area index (LAI) and SPAD value serve as key physiological parameters. SPAD is an indicator of leaf chlorophyll content, and together with LAI, they reflect canopy structure, photosynthetic capacity, and nutritional status (Zaji et al., 2022; Zhang et al., 2024). However, traditional manual methods for measuring these indicators are often labor-intensive, destructive, and unsuitable for the large-scale, high-frequency monitoring required in modern precision agriculture (Chen et al., 2025).

Recent advancements in unmanned aerial vehicle (UAV)-based multispectral remote sensing have emerged as a powerful solution for crop monitoring, offering high-resolution and flexible acquisition of canopy spectral information (Qiao, Gao, et al., 2022). Equipped with multispectral sensors, UAVs capture reflectance data across visible, red-edge, and near-infrared bands—spectral regions closely correlated with crop biochemical and biophysical traits (Gao et al., 2022). From these spectral bands, various vegetation indices (VIs) are derived, including the normalized difference vegetation index (NDVI), enhanced vegetation index (EVI), and soil-adjusted vegetation index (SAVI). These indices allow quantitative assessments of canopy growth, biomass, and chlorophyll content. Each index possesses unique advantages tailored to specific application scenarios (Zeng et al., 2022). However, conventional VIs tend to saturate under dense canopies (typically when LAI exceeds 2.5), diminishing their sensitivity to LAI and SPAD values (Han et al., 2025; Wan et al., 2020). Previous research has attempted to mitigate this challenge by incorporating texture features (TFs) extracted from VI images. However, these handcrafted descriptors still show limited sensitivity when the canopy becomes highly closed and structural variations are subtle (Qiao, Zhao, et al., 2022). This highlights a pressing need for more expressive image features capable of capturing nuanced LAI differences.

Machine learning techniques, including random forest (RF), support vector machines (SVM), and extreme gradient boosting (XGBoost), have been employed to model nonlinear relationships between VIs and crop parameters, enhancing prediction accuracy (Qi et al., 2025). However, "point-based" approaches assign a single LAI or SPAD measurement to each plot and utilize averaged VI values. As a result, they fail to fully exploit the rich spatial-spectral information present in UAV imagery, which constrains overall model performance. In recent years, deep learning—particularly convolutional neural networks (CNNs)—has shown promise in end-to-end extraction of spectral-spatial features, demonstrating superior accuracy in LAI and SPAD estimation. Notable applications include maize LAI estimation from UAV VI imagery (Qiao, Zhao, et al., 2022) and the integration of hyperspectral data with radiative transfer models (RTMs) for LAI and chlorophyll prediction (Yue et al., 2024). Early-stage LAI estimation in winter wheat has also been performed using time-series field images (Li et al., 2021). Despite these advances, the broader adoption of deep learning in this domain is impeded by several challenges. First, limited labeled data can lead to overfitting (Jia et al., 2021). Second, conventional dataset partitioning strategies may fail to provide representative learning across growth stages and are susceptible to noise (Turney, 1994). Third, general-purpose CNN architectures such as ResNet and VGG, pretrained on RGB images, often exhibit poor generalization to multispectral or VI-based regression tasks due to substantial domain gaps (Penatti et al., 2015; Wang et al., 2023). Finally, their considerable computational and memory demands hinder deployment on resource-constrained platforms such as UAVs (Zidi & Abdelkrim, 2024).

Recent innovations in self-supervised learning (SSL) and attention mechanisms present promising solutions to these challenges. SSL methods, such as VICReg, leverage large unlabeled datasets to reliance on labeled samples (Bardes et al., 2022), while attention modules effectively highlight critical spectral channels and spatial regions (Hu et al., 2018). However, there is a notable gap in the adaptation of these strategies specifically for addressing VI saturation in dense wheat canopies or for developing lightweight models suitable for UAV deployment.

To bridge these gaps, this study proposes the Multi-Channel Vegetation Indices Saturation Aware Net (MCVI-SANet), a vision model specifically designed for high-resolution vegetation index imagery. The model focuses on estimating LAI and SPAD under limited-sample conditions in winter wheat. The key innovations of this study are as follows.

(1) A lightweight semi-supervised network named MCVI-SANet is proposed for

estimating winter wheat LAI and SPAD under limited-sample conditions. It employs VICReg-based self-supervised pretraining to enhance generalization. Additionally, it utilizes a phenology-informed data partitioning strategy to ensure representative learning across growth stages, even under high-noise and limited-sample conditions.

(2) The Vegetation Indices Saturation Aware Block (VI-SABlock) is designed to mitigate VI saturation under dense canopy conditions. It adaptively fuses multi-channel VI statistics to enhance informative spectral-spatial features. It also jointly encodes VI priors with spatial-structural information from multi-channel VI imagery, enabling more accurate characterization of canopy conditions and mitigating the impact of VI saturation.

(3) Extensive experiments demonstrate that MCVI-SANet achieves state-of-the-art accuracy and high efficiency. It outperforms both traditional machine learning and existing deep learning methods, while maintaining a minimal number of model parameters.

This study highlights the potential of integrating semi-supervised learning with agronomic principles to enhance remote sensing-based precision agriculture, thereby contributing to more informed decision-making in crop management practices.

## 2. Methodology

*2.1. Data acquisition*

*2.1.1. Study area*

The study was conducted in the experimental field of the Science and Innovation Base of Shandong Agricultural University, Tai'an, Shandong Province, China (36°9'30"N, 117°9'13"E). The region features a warm-temperate, semi-humid continental monsoon climate, with an average annual temperature of 12.9°C, 2,627.1 h of sunshine, an 697 mm of precipitation. The winter wheat was sown on October 25, 2024, and harvested on June 15, 2025, with standard field management applied throughout the growing season. A total of 60 sampling plots (3 m×3 m each, 9 m$^2$ per plot) were evenly distributed across the field, with approximately 1 meter of spatial isolation boundary separating each plot to reduce spatial autocorrelation between different sampling areas as much as possible (Fig. 1). UAV-based multispectral imagery and crop growth parameters were collected from these plots.

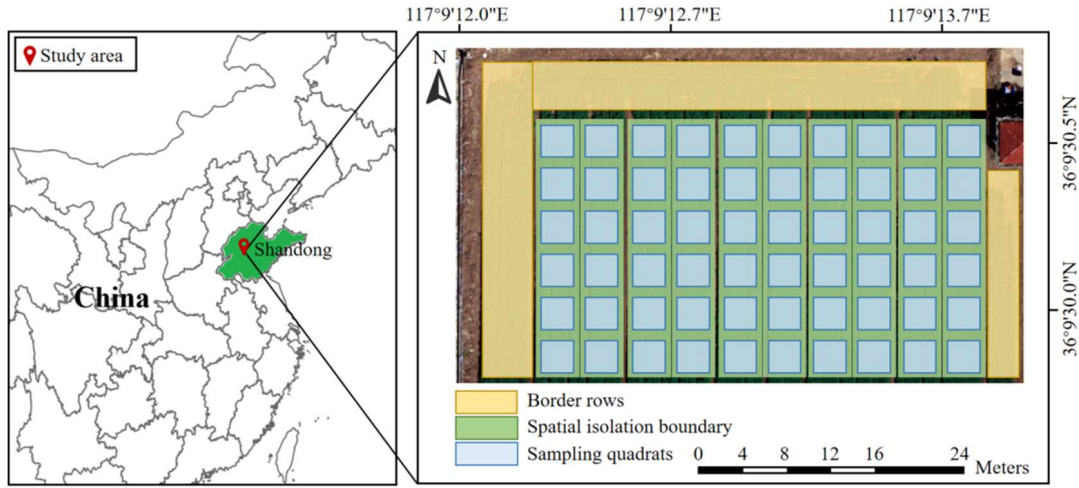

**Fig. 1.** Experimental design of the study area.

*2.1.2. UAV multispectral imaging*

Multispectral imagery was acquired using a DJI M300 RTK UAV (DJI Inc., China) equipped with a RedEdge-P multispectral camera (MicaSense Inc., USA). All flights were conducted under clear, low-cloud, and windless conditions between 10:00 and 14:00 to ensure accurate capture of canopy spectral and structural information. Prior to each flight, a ground radiometric calibration panel was imaged to support subsequent radiometric correction. The multispectral camera is integrated with a downwelling light sensor, which synchronously records solar irradiance to correct for illumination variations.

Flight parameters were configured to ensure geometric accuracy, with 80% forward and side overlap, a flight altitude of 15 m above ground, and the use of Ground Control Points (GCPs). The camera simultaneously acquired five spectral bands: red (R), green (G), blue (B), red edge (RE), and near-infrared (NIR) (Table 1). Each flight generated approximately 1,500 images. UAV flights were performed at key wheat growth stages: jointing stage (April 9 and 14, 2025), booting stage (April 23), heading stage (April 28), flowering stage (May 11), and grain-filling stage (May 20), resulting in a total of ~9,200 valid multispectral images.

**Table 1**
Shooting parameters of the multispectral camera.

| Band | Center wavelength (nm) | Bandwidth (nm) | Resolution |
| --- | --- | --- | --- |
| Red | 668 nm | 14 nm | 1456×1088 |
| Green | 560 nm | 27 nm | 1456×1088 |
| Blue | 475 nm | 32 nm | 1456×1088 |

| Red Edge | 717 nm | 12 nm | 1456×1088 |
| Near Infrared | 842 nm | 57 nm | 1456×1088 |

**Table 2**
Acquisition dates of winter wheat growth parameter measurements.

| Growth stages | Dates | Simple number |
|---|---|---|
| Jointing stage | April 9, 2025 | 30 |
| | April 14, 2025 | 60 |
| Booting stage | April 23, 2025 | 60 |
| Heading stage | April 28, 2025 | 60 |
| Flowering stage | May 11, 2025 | 60 |
| Grain filling stage | May 20, 2025 | 60 |

*2.1.3. Ground crop growth measurements*

The key growth parameters of winter wheat measured in this study not only included LAI and SPAD, but also additional data on vegetation height (VH). Standardized multi-point sampling procedures were employed to ensure accuracy and representativeness:

(1) LAI: Measured with an AccuPAR LP-80 canopy analyzer (METER Group Inc., USA) at five points per plot, with the arithmetic mean taken as the plot LAI.

(2) SPAD: Measured with a SPAD-502 (Konica Minolta Inc., Japan) chlorophyll meter on four representative plants per plot, and averaged to obtain the plot SPAD value.

(3) VH: Measured with a tape at the center of each plot for high, medium, and low representative plants, with the average value recorded as the plot VH.

Ground measurements were conducted simultaneously with UAV flights. In total, 330 data sets of growth parameters were collected across five growth stages (Table 2); due to operational constraints, only 30 plots were sampled on April 9. Boxplots of LAI, SPAD, and VH across all stages are shown in Fig. 2, with J1 and J2 representing the first and second jointing stage measurements, B for booting, H for heading, F for flowering, and G for grain-filling stages.

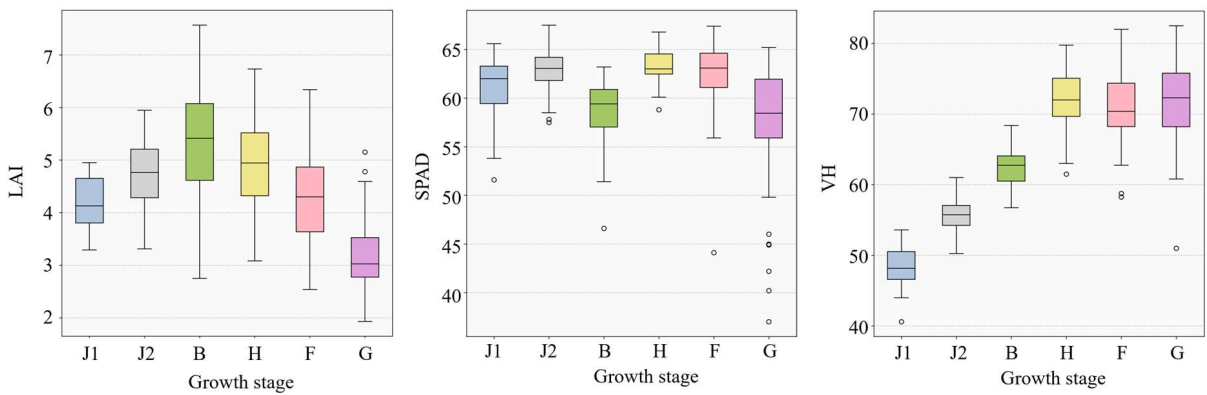

(a) LAI (m$^2$/m$^2$)     (b) SPAD (SPAD-502 reading)     (c) VH (cm)

**Fig. 2.** Boxplots of measured winter wheat growth parameters at key phenological stages.

**Table 3**

Results of UAV multispectral image processing.

| Type | Bands | Image size | Quantity | Corresponding Labels |
| --- | --- | --- | --- | --- |
| Labeled Images | Red, Green, Blue, Red edge, NIR | 192×192 | 330 per band | LAI, SPAD, VH |
| Unlabeled Images | Red, Green, Blue, Red edge, NIR | 192×192 | 2700 per band | No labels |

*2.2. Data processing*

*2.2.1. UAV multispectral image processing*

Post-processing of UAV multispectral imagery was conducted using Pix4Dmapper software (Pix4D S.A., Switzerland), including radiometric calibration, geometric correction, and orthomosaic generation. A complete five-band multispectral orthomosaic was produced for each acquisition date. Based on these orthomosaics, image tiles corresponding to each sampling plot were clipped in ArcGIS (Esri, USA) with a spatial size of 192×192 pixels per plot, resulting in 330 labeled samples across six acquisition dates. Each sample was paired with measured LAI, SPAD, and VH values (Table 3).

In addition, multispectral images were also collected from an extra southern area managed under the same practices, and all images were uniformly segmented into unlabeled five-band image tiles of the same size, generating 2,700 unlabeled images for self-supervised pretraining (Table 3).

**Table 4**

Vegetation indices used in this study.

| Vegetation Index | Formula | Reference |
| --- | --- | --- |
| Chlorophyll index with green (CIgreen) | $\dfrac{NIR}{G} - 1$ | (Bagheri & Kafashan, 2025) |
| Difference vegetation index (DVI) | $NIR - R$ | (Li et al., 2023) |
| Enhanced vegetation index (EVI) | $\dfrac{2.5(NIR - R)}{NIR + 6R - 7.5B + 1}$ | (Bagheri & Kafashan, 2025) |
| Green normalized difference vegetation index (GNDVI) | $\dfrac{NIR - G}{NIR + G}$ | (Ochiai et al., 2024) |
| Modified chlorophyll absorption in reflectance | $\dfrac{RE(RE - R - 0.2(RE - G))}{R}$ | (Joshi et al., 2024) |

| | | | |
|---|---|---|---|
| index (MCARI) | | | |
| Normalized difference red edge index (NDRE) | $\dfrac{NIR - RE}{NIR + RE}$ | (Bagheri & Kafashan, 2025) | |
| Normalized difference vegetation Index (NDVI) | $\dfrac{NIR - R}{NIR + R}$ | (Bagheri & Kafashan, 2025) | |
| Optimized soil-adjusted vegetation index (OSAVI) | $\dfrac{1.16(NIR - R)}{NIR + R + 0.16}$ | (Bagheri & Kafashan, 2025) | |
| Ratio vegetation index (RVI) | $\dfrac{NIR}{R}$ | (Ma et al., 2024) | |
| Soil-adjusted vegetation index (SAVI) | $\dfrac{1.5(NIR - R)}{NIR + R + 0.5}$ | (Ochiai et al., 2024) | |
| Visible atmospherically resistant index (VARI) | $\dfrac{G - R}{G + R - B}$ | (Ochiai et al., 2024) | |

Note: R=red，G=green，B=blue，RE=red edge，NIR=near-infrared

*2.2.2. VI images generation and hand-crafted features extraction*

To construct a representative feature set, 11 vegetation indices widely validated in crop monitoring were selected based on previous studies (Table 4). These indices were derived from the red, green, blue, red-edge, and near-infrared spectral bands, thereby integrating complementary spectral information across multiple wavelengths.

For each of the 330 labeled multispectral images, 11 corresponding VI images were generated. The mean pixel value of each VI image within a plot was then calculated and used as the numerical VI feature, forming the basis of the supervised modeling dataset. In addition, based on the findings of Qiao, Zhao, et al. (2022), six TFs derived from NDRE were selected, which exhibits relatively low saturation effects during periods of dense canopy cover. These features include mean, standard deviation, smoothness, third moment, uniformity, and entropy. Their calculation formulas are provided in Equations (1)–(6).

$$Mean = \frac{1}{N}\sum_{i=1}^{N} I_i \qquad (1)$$

$$Standard\text{-}deviation = \sqrt{\frac{1}{N}\sum_{i=1}^{N}(I_i - \mu)^2} \qquad (2)$$

$$Smoothness = 1 - \frac{1}{1 + \sigma^2} \qquad (3)$$

$$Third\text{-}moment = \frac{1}{N}\sum_{i=1}^{N}(I_i - \mu)^3 \tag{4}$$

$$Uniformity = \sum_{k=1}^{K} p_k^2 \tag{5}$$

$$Entropy = -\sum_{k=1}^{K} p_k \ln(p_k) \tag{6}$$

Where $I_i$ denotes the intensity value of the $i$-th pixel in the image, $N$ represents the total number of pixels, $p_k$ is the normalized probability associated with the $k$-th gray-level bin obtained from the image histogram, and $K$ is the total number of histogram bins.

### 2.2.3. Dataset partitioning method

To ensure balanced datasets under limited-sample conditions and improve evaluation stability, a stratified sampling strategy combining VH-guided K-Means clustering and a majority-voting mechanism was adopted. Drawing inspiration from Zeng et al. (2014), who proposed the use of auxiliary vegetation parameters and stratified sampling schemes for representative sampling of remotely sensed LAI products over heterogeneous surfaces, VH is incorporated as a non-model covariate and additional growth feature that increases consistently with phenological development. This enables improved representativeness of training, validation, and test sets across both feature and phenological dimensions. The approach mitigates temporal bias inherent in LAI- and SPAD-based clustering and ensures uniform coverage of growth stages. It aims to reduces overfitting under high-noise, small-sample, and complex-model scenarios. The dataset partitioning algorithm was developed as follows.

Let the original dataset $D$ contain $|D|$ samples, each characterized by three growth parameters (LAI, SPAD, VH), forming a feature matrix $F \in \mathbf{R}^{|D| \times 3}$. To eliminate scale differences, feature normalization is first performed:

$$F_{norm} = \frac{F - \mu}{\sigma} \tag{7}$$

Where $\mu$ and $\sigma$ denote the mean and standard deviation of each feature, respectively. The algorithm then applies K-Means clustering with $K=10$ to capture the distributional structure of samples in the growth parameter space. To reduce sensitivity to random initialization, clustering is repeated $N > 5$ times, producing a label matrix $L \in \mathbf{Z}^{N \times |D|}$. For each sample $j$, its final cluster label $c_j$ is determined by majority voting to obtain a more stable and reliable cluster assignment:

$$c_j = \arg\max_{k \in \{1,\ldots,K\}} \sum_{i=1}^{N} \mathrm{I}(L_{ij} = k) \tag{8}$$

Where $\mathrm{I}(\cdot)$ is the indicator function. Within each final cluster $I_c$, stratified sampling is then performed according to global ratios $R_{\text{train}} : R_{\text{val}} : R_{\text{test}} = 9 : 1 : 1$. The number of samples assigned to each subset is computed as:

$$n_{\text{train}}^{(c)} = \lfloor |I_c| \cdot R_{\text{train}} \rfloor \tag{9}$$

$$n_{\text{val}}^{(c)} = \lfloor |I_c| \cdot R_{\text{val}} \rfloor \tag{10}$$

$$n_{\text{test}}^{(c)} = |I_c| - n_{\text{train}}^{(c)} - n_{\text{val}}^{(c)} \tag{11}$$

The mean and standard deviation of the training set, validation set and test set after division are shown in Table 5. Furthermore, the average maximum mean discrepancy (MMD) between the data subsets was $1.3 \times 10^{-3}$, and the Jensen-Shannon (JS) was 0.557, and the mean coefficient of variation (CV) across different growth stages was 0.220. In contrast, using only LAI and SPAD yielded higher distribution shifts (MMD = $2.6 \times 10^{-3}$, JS = 0.563, CV = 0.272), indicating that incorporating VH effectively reduces inter-stage distribution discrepancies and enhances the representativeness and stability of the partitioned subsets.

**Table 5**
Statistics of the training set, test set and validation set.

| Parameter | Mean | | | Standard deviation | | |
|---|---|---|---|---|---|---|
| | Train set | Valid set | Test set | Train set | Valid set | Test set |
| LAI | 4.47 | 4.46 | 4.37 | 1.08 | 1.07 | 1.13 |
| SPAD | 61.06 | 60.10 | 60.63 | 4.41 | 3.37 | 4.39 |

*2.3. Construction of VI-based estimation models using traditional machine learning methods*

Traditional estimation models for winter wheat growth parameters are typically constructed based on VIs combined with machine learning algorithms to establish quantitative relationships. Following previous research, four representative algorithms were selected to develop baseline estimation models:

(1) Partial least squares regression (PLSR), a linear method capable of handling multicollinearity among predictors;

(2) Random forest regression (RFR) and XGBoost — nonlinear ensemble learning methods that capture complex feature interactions;

(3) Support vector regression (SVR) — suitable for high-dimensional and nonlinear regression tasks.

To ensure fairness and reliability in comparative experiments, all machine learning models were optimized using the Optuna (Akiba et al., 2019) framework, which performs automated bayesian optimization based on the tree-structured parzen estimator algorithm to efficiently explore the parameter space.

The specific hyperparameter search ranges for each model are listed in Table 6, ensuring optimal configurations under consistent conditions while significantly improving parameter tuning efficiency.

**Table 6**
Machine learning model parameters and search ranges.

| Model | Hyperparameter | Search space | Description |
|---|---|---|---|
| PLSR | n_components | [1, min(11, 20)] | Number of latent components |
| RFR | n_estimators | [5, 150] | Number of trees in the forest |
|  | max_depth | [1, 20] | Maximum depth of each tree |
|  | min_samples_split | [2, 10] | Minimum samples to split internal nodes |
|  | min_samples_leaf | [1, 5] | Minimum samples at a leaf node |
|  | max_features | ['sqrt', 'log2', None] | Maximum features considered for splitting |
| XGBoost | n_estimators | [5, 150] | Number of trees |
|  | max_depth | [1, 20] | Maximum depth per tree |
|  | learning_rate | [0.01, 0.3] | Learning rate |
|  | subsample | [0.5, 1] | Fraction of samples for training each tree |
|  | colsample_bytree | [0.5, 1] | Fraction of features for training each tree |
|  | reg_alpha | [0, 10] | L1 regularization coefficient |
|  | reg_lambda | [0, 10] | L2 regularization coefficient |
| SVR | C | [0.1, 100] | Regularization parameter |
|  | epsilon | [0.001, 1] | Epsilon-insensitive zone |
|  | gamma | ['scale, 'auto'] | Kernel coefficient |

*2.4. Construction of MCVI-SANet*

This study proposed a model named MCVI-SANet, designed for accurate estimation of LAI and SPAD under limited-sample conditions. The overall architecture is illustrated in Fig. 3. MCVI-SANet leverages a lightweight architectural design to enable highly efficient estimation of LAI and SPAD, achieving substantial reductions in computational cost without compromising predictive accuracy.

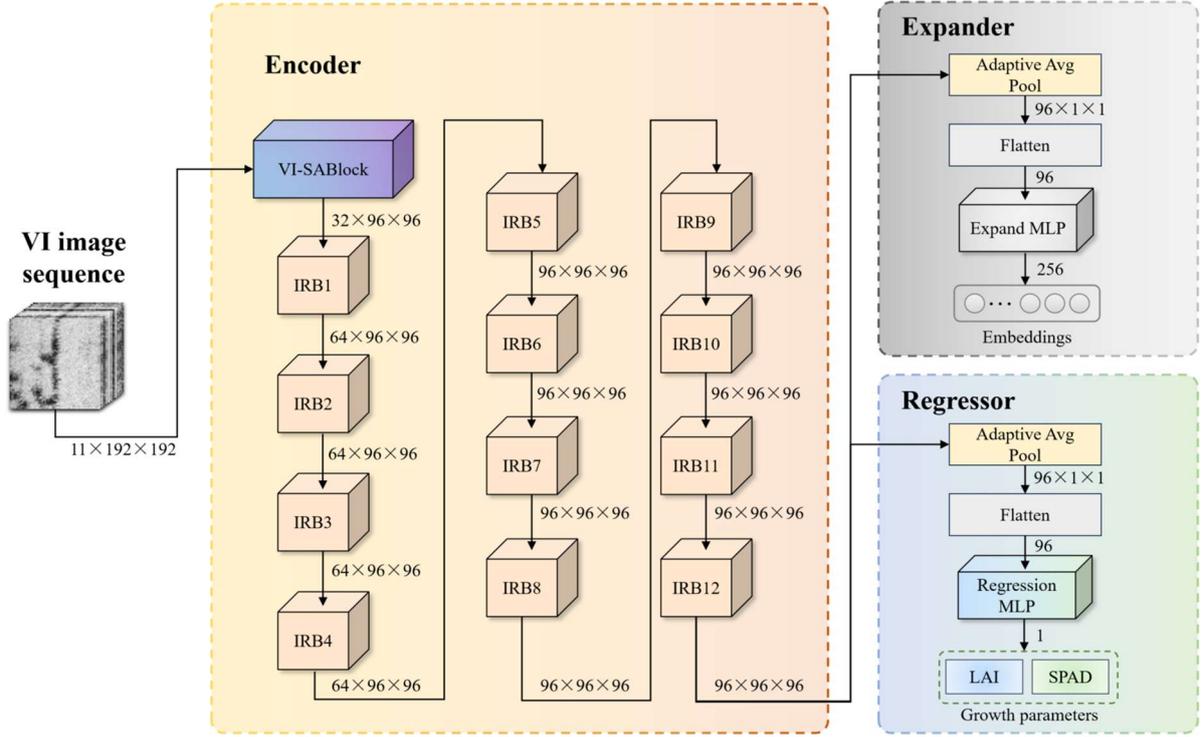

**Fig. 3.** Overall architecture of the proposed MCVI-SANet model.

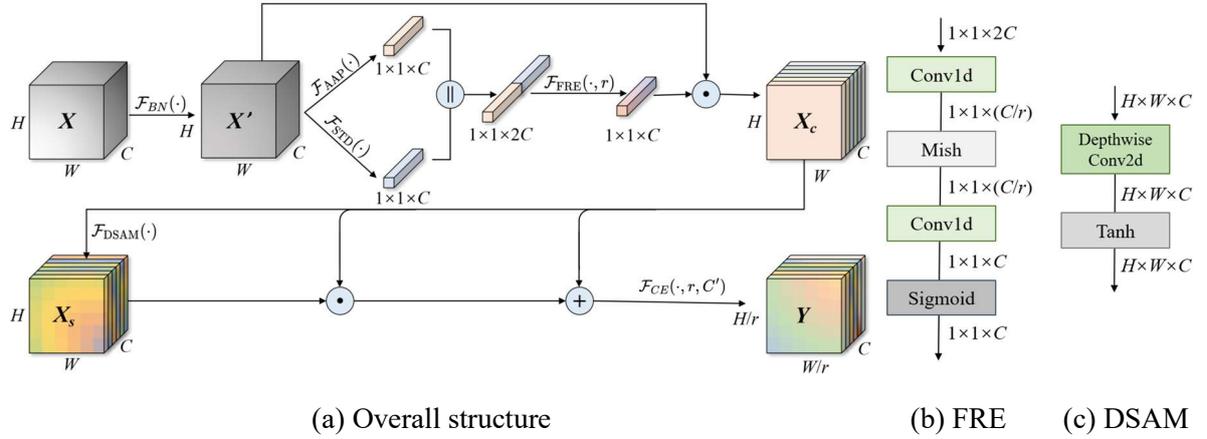

(a) Overall structure          (b) FRE     (c) DSAM

**Fig. 4.** Architecture of the VI-SABlock.

### 2.4.1. VI-SABlock

To alleviate the spectral saturation effect commonly observed in dense wheat canopies, this study proposed the VI-SABlock (Fig. 4 (a)) as the first layer of the network for front-end adaptive feature enhancement. This module incorporates a dual-attention mechanism based on the statistical properties of multi-channel VIs.

A batch normalization (BN) layer is introduced at the beginning of the block to stabilize the distribution of multi-channel inputs and facilitate smoother optimization. Given the scale discrepancies among VIs, such as NDVI ranging within [-1,1] and RVI spanning [0,∞), it

helps normalize heterogeneous feature distributions before attention modeling:

$$X' = F_{BN}(X) \in \mathbf{R}^{C \times H \times W} \tag{12}$$

After this, both adaptive average pooling (AAP) and standard deviation (STD) descriptors were utilized to capture and fuse multi-index information adaptively:

$$\boldsymbol{\mu} = F_{AAP}(X') = \frac{1}{HW}\sum_{i=1}^{H}\sum_{j=1}^{W} X'_{c,i,j} \in \mathbf{R}^{C \times 1 \times 1} \tag{13}$$

$$\boldsymbol{\sigma} = \sqrt{F_{AAP}\left((X')^2\right) - \boldsymbol{\mu}^2 + \epsilon} \in \mathbf{R}^{C \times 1 \times 1} \tag{14}$$

Where $H$ and $W$ denote the height and width of the feature map, and $\epsilon = 10^{-10}$ ensures numerical stability. Because different VIs depend on distinct spectral bands, their responses vary under changing canopy conditions (Zeng et al., 2022). As canopy density increases, common indices such as NDVI tend to saturate, weakening the sensitivity of GAP to biomass changes. In contrast, the STD of VI images retains structural information (e.g., leaf angle distribution, light-shadow patterns, and micro-stress patches), which remains valuable even under saturation. Therefore, not only GAP, but also STD was applied as the feature descriptor. Although BN standardizes multi-channel inputs across the batch, GAP and STD remain effective since they capture spatial statistics within individual samples, and BN's linear channel-wise transformation preserves the relative structural variations and global intensity patterns.

Then, The descriptors are concatenated as:

$$C = \text{Concat}(\boldsymbol{\mu}, \boldsymbol{\sigma}) \in \mathbf{R}^{2C \times 1 \times 1} \tag{15}$$

The concatenated statistics are then processed by a feature recalibration excitation (FRE) network (Fig.4 (b)) to learn nonlinear dependencies between statistics and adaptively reweight the VI channels:

$$A_{channel} = F_{FRE}(C, r) = \text{Sigmoid}\left(W_2 \text{Mish}(W_1[\boldsymbol{\mu}, \boldsymbol{\sigma}])\right) \in \mathbf{R}^{C \times 1 \times 1} \tag{16}$$

Where $W_1 \in \mathbf{R}^{(C/r) \times 2C}$ and $W_2 \in \mathbf{R}^{C \times (C/r)}$ are 1×1 convolutional layers. Considering that the values of some VIs fall within the range of negative numbers, the Mish activation function is used instead of ReLU to avoid truncating negative values and enhance smooth gradient flow:

$$\text{Mish}(z) = z \cdot \tanh(\ln(1 + e^z)) \tag{17}$$

Compared to ReLU, Mish is continuously differentiable across the entire range with smoother gradient changes. This enables the model to respond more accurately to negative inputs during training, thereby accelerating convergence and enhancing feature representation.

The reweighted channel attention is applied to the input features to enhance the key VI channels and suppress the non-key channels:

$$X_c = X' \odot A_{channel} \in \mathbf{R}^{C \times H \times W} \tag{18}$$

To further enhance spatial discriminability under saturation, a depthwise spatial attention module (DSAM) (Fig.4 (c)) is introduced. Unlike CBAM, which relies on average and max pooling, DSAM employs depthwise convolutions to learn fine-grained spatial weights. When VIs enter the saturation region, subtle spatial variations in canopy reflectance become the key cues for distinguishing different growth states. At this stage, average pooling and max pooling exhibit inherent limitations: max pooling is highly sensitive to noise and outliers, whereas average pooling tends to smooth out fine structural differences within the canopy (Zhao & Zhang, 2024). In contrast, the learnable depthwise filters can scan feature maps channel by channel through local receptive fields, allowing more precise capture of weak textural patterns and localized structural variations. Meanwhile, it effectively suppresses saturated regions across channels and spatial positions.

Therefore, this design provides a unique advantage for feature enhancement under VI saturation conditions. The computation of DSAM is defined as follows:

$$A_{spatial} = F_{\text{DSAM}}(X_c) = \tanh(G_{depth} X_c) \in \mathbf{R}^{C \times H \times W} \tag{19}$$

where $G_{depth}$ is a 3×3 depthwise convolution and $\tanh(\cdot)$ provides zero-centered outputs within [-1,1] enabling bidirectional modulation—positive activation regions enhance feature responses that are sensitive to canopy structural variations, whereas negative regions suppress ineffective features potentially caused by saturation. In addition, compared with the sigmoid, the zero-centered output of tanh helps reduce weight bias during training, thereby improving training stability and accelerating convergence (Glorot & Bengio, 2010).

The final spatial attention is implemented in a residual manner:

$$X_s = X_c \odot (1 + A_{spatial}) \in \mathbf{R}^{C \times H \times W} \tag{20}$$

This design preserves the original feature distribution while selectively strengthening or suppressing key spatial regions, thereby preventing additional distribution shifts under saturation conditions.

Finally, an expansion layer increases the channel dimension to $C^{'}$ for enrich features representation while downsampling the spatial resolution by a factor of $r$ to reduce computation, facilitating efficient downstream processing:

$$Y = F_{\text{CE}}(X_s, r, C^{'}) \in \mathbf{R}^{C^{'} \times (H/r) \times (W/r)} \tag{21}$$

*2.4.2. Lightweight backbone network*

The parameters of the lightweight backbone network are summarized in Table 7. The backbone of MCVI-SANet is built upon Inverted Residual Blocks (IRBs) (Sandler et al., 2018). As shown in Fig. 3, 12 IRBs are stacked to ensure efficient feature extraction under lightweight constraints. All IRBs are configured with stride=1 and residual connections to preserve the fine-grained canopy structures that are critical for mitigating VI saturation. This identity-preserving design also facilitates stable gradient propagation, ultimately producing a 96×96 feature map.

**Table 7**
Parameters of lightweight backbone.

| Block name | Operation | Output shape |
| --- | --- | --- |
| IRB1 | Inverted Residual (t=1, c=64, s=1) | (N, 64, 96, 96) |
| IRB2 | Inverted Residual (t=1, c=64, s=1) | (N, 64, 96, 96) |
| IRB3 | Inverted Residual (t=1, c=64, s=1) | (N, 64, 96, 96) |
| IRB4 | Inverted Residual (t=1, c=64, s=1) | (N, 64, 96, 96) |
| IRB5 | Inverted Residual (t=1, c=96, s=1) | (N, 96, 96, 96) |
| IRB6 | Inverted Residual (t=1, c=96, s=1) | (N, 96, 96, 96) |
| IRB7 | Inverted Residual (t=1, c=96, s=1) | (N, 96, 96, 96) |
| IRB8 | Inverted Residual (t=1, c=96, s=1) | (N, 96, 96, 96) |
| IRB9 | Inverted Residual (t=1, c=96, s=1) | (N, 96, 96, 96) |
| IRB10 | Inverted Residual (t=1, c=96, s=1) | (N, 96, 96, 96) |
| IRB11 | Inverted Residual (t=1, c=96, s=1) | (N, 96, 96, 96) |
| IRB12 | Inverted Residual (t=1, c=96, s=1) | (N, 96, 96, 96) |

[1]t: the expansion factor of the IRB.
[2]c: the number of output feature channels produced by the IRB.
[3]s: the stride of the depthwise convolution in the IRB.
[4]N: the batch size.

*2.4.3. Semi-supervised training strategy*

A two-stage semi-supervised learning strategy based on VICReg was adopted to address the limited availability of labeled samples and enhance the model's generalization capability. This strategy first performs self-supervised pretraining on unlabeled multi-channel VI imagery to enable the encoder to learn general-purpose feature representations. The pretrained encoder is then transferred and fine-tuned on the labeled samples for the final regression-based estimation tasks.

Specifically, for a batch of input images, VICReg applies two independent random augmentations, *t* and *t'*, generating two augmented views *X* and *X'*. These are passed through the encoder to obtain representations *Y* and *Y'*, which are then expanded by the expander into high-dimensional embeddings *Z* and *Z'*.

The learning process is constrained by three independent regularization terms: invariance, variance, and covariance (Equations (22)–(24)). The invariance loss minimizes the euclidean distance between the embeddings of different augmented views of the same image, enforcing semantic consistency. The variance loss ensures that each embedding dimension maintains a minimum variance within the batch, preventing feature collapse and preserving diversity. The covariance loss penalizes the off-diagonal elements of the covariance matrix of embeddings, encouraging the model to learn non-redundant representations.

$$L_{\text{sim}} = \frac{1}{B} \sum_{i=1}^{B} \| Z_i^{(1)} - Z_i^{(2)} \|_2^2 \tag{22}$$

$$L_{\text{var}} = \frac{1}{D} \sum_{d=1}^{D} \max\left(0, \gamma - \sqrt{\text{Var}(Z_d) + \epsilon}\right) \tag{23}$$

$$L_{\text{cov}} = \frac{1}{D} \sum_{i \neq j} [C(Z)]_{i,j}^2 \tag{24}$$

$$L = \lambda_{\text{sim}} L_{\text{sim}} + \lambda_{\text{var}} L_{\text{var}} + \lambda_{\text{cov}} L_{\text{cov}} \tag{25}$$

Here, $\text{Var}(Z_d)$ denotes the batch variance of the d-th feature dimension, and $C(Z)$ represents the covariance matrix of batch features. The batch size *B* was set to 100 to balance the stability of variance estimation and the computational resource usage. The remaining hyper-parameters followed those of the original VICReg implementation: $\lambda_{\text{sim}} = 25.0$, $\lambda_{\text{var}} = 25.0$, $\lambda_{\text{cov}} = 1.0$ (Equations (25)), $\epsilon = 1 \times 10^{-4}$, $D = 256$, and $\gamma = 1$.

During the self-supervised pretraining stage, the encoder and expander are jointly trained on 2,700 unlabeled multi-channel images. Each image undergoes random horizontal/vertical flipping and rotation by multiples of 90°, ensuring augmentation diversity while avoiding interpolation artifacts, boundary padding, or spectral distortion that could disrupt the correspondence between imagery and growth parameters.

The projector consists of two fully connected layers with BN and ReLU activation, both having an output dimension of 256, mapping the encoder output into a 256-dimensional latent space. After pretraining, the projection head is removed and the encoder weights are frozen, followed by attaching the regressor for downstream tasks. The regressor also comprises two fully connected layers, with the first projecting features into a 32-dimensional latent space and the second generating the final LAI or SPAD estimate.

## 3. Experimental setup

*3.1. Experimental details*

All experiments were conducted on a workstation equipped with an Intel Core i9-14900K CPU and dual NVIDIA GeForce RTX 3090 GPUs (each with 24 GB of VRAM). The software environment was based on Windows 11 Professional, and all codes were developed using Python 3.11.

Considering the differences in model architectures, differentiated training strategies were adopted. For traditional machine learning models, hyperparameter optimization was performed automatically using the Optuna framework for efficient search. For the self-supervised VICReg pretraining, the initial learning rate was set to $1\times10^{-3}$, with 500 epochs and a batch size of 100. For supervised training, the initial learning rate of LAI estimation was set to $5\times10^{-4}$, and $5\times10^{-5}$ for SPAD estimation, both trained for 200 epochs with a batch size of 32, and the early stopping strategy was employed to prevent overfitting. Moreover, LogME (You et al., 2021) was incorporated as an auxiliary criterion to rank and select the pre-trained self-supervised models most suitable for the downstream task.

*3.2. Model accuracy evaluation*

The performance of all models in estimating LAI and SPAD was quantitatively evaluated using $R^2$ and RMSE. The evaluation metrics are defined as follows:

$$R^2 = 1 - \frac{\sum_{i=1}^{n}(y_i - \hat{y}_i)^2}{\sum_{i=1}^{n}(y_i - \bar{y})^2} \quad (26)$$

$$RMSE = \sqrt{\frac{1}{n}\sum_{i=1}^{n}(y_i - \hat{y}_i)^2} \quad (27)$$

Where $n$ denotes the total number of samples, $y_i$ is the ground truth value of the $i$-th sample,

$\hat{y}_i$ represents the corresponding predicted value, and $\bar{y}$ is the mean of all observed values.

## 4. Results

*4.1. The self-supervised representation learning*

The loss curves during VICReg pretraining phase are illustrated in Fig. 5, exhibiting trends that align with theoretical expectations.

Specifically, the similarity loss, variance loss, and total loss demonstrated a rapid decline during the initial training phase, followed by a gradual decline in convergence gradients until stabilization was achieved. The covariance loss initially rose sharply and then gradually decreased, mainly because the high-weight similarity loss initially forces the two augmented views to align closely, temporarily inducing stronger inter-dimensional correlations. As consistency learning stabilizes, the covariance loss decreases and converges, indicating that the model has effectively learned stable, non-collapsed representations.

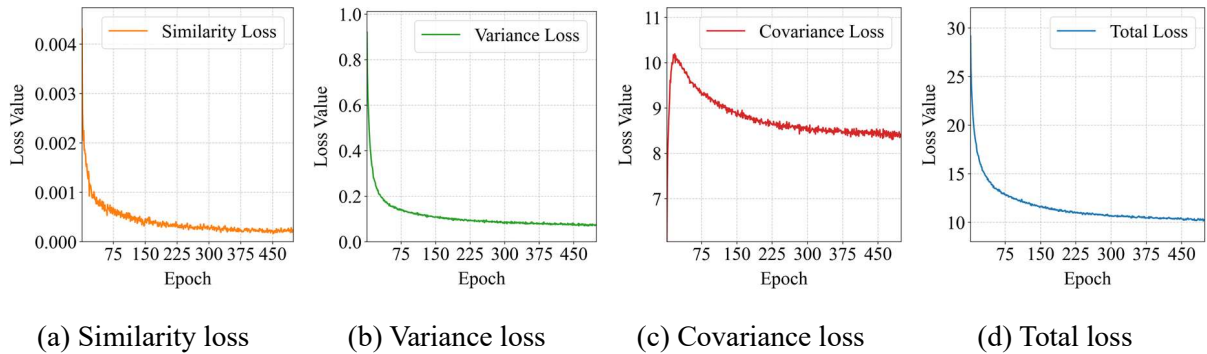

(a) Similarity loss    (b) Variance loss    (c) Covariance loss    (d) Total loss

**Fig. 5.** Loss curves of VICReg self-supervised pretraining.

*4.2. Comparative experiments of LAI and SPAD estimate models*

To evaluate the performance of different models in LAI and SPAD estimation tasks, in addition to traditional machine learning models, 7 mainstream deep learning baselines were selected for comparison, including EfficientNet B0, EfficientNet V2-S, MobileNet V2, MobileNet V3, ResNet18, ResNet50, and ShuffleNet V2. To minimize the influence of randomness, the experiments were conducted 10 times independently, and the final $R^2$ and RMSE values were obtained by averaging the outcomes across the 10 runs.

The comparative results are summarized in Table 8. For visualization purposes, the best-performing model within each model category was selected, and its top-performing run was used to generate the predicted versus true values scatter plots on the test set (Fig. 6).

**Table 8**

Comparative results of different models on the test dataset.

| Model | | LAI | | SPAD | |
|---|---|---|---|---|---|
| | | $R^2$ | RMSE | $R^2$ | RMSE |
| Traditional machine learning models | PLSR (VIs) | 0.5103 | 0.7756 | **0.6329** | **2.6143** |
| | PLSR (VIs+TFs) | 0.5206 | 0.7673 | 0.5881 | 2.7694 |
| | RFR (VIs) | 0.6677 | 0.6388 | 0.6326 | 2.6151 |
| | RFR (VIs+TFs) | 0.6835 | 0.6230 | 0.6327 | 2.6139 |
| | SVR (VIs) | 0.6855 | 0.6216 | 0.5768 | 2.8069 |
| | SVR (VIs+TF) | **0.7278** | **0.5782** | 0.4739 | 3.1297 |
| | XGBoost (VIs) | 0.6265 | 0.6770 | 0.6256 | 2.6378 |
| | XGBoost (VIs+TFs) | 0.6920 | 0.6149 | 0.6089 | 2.6967 |
| Baseline deep learning models | EfficientNet B0 | 0.6603 | 0.6445 | 0.4552 | 3.1520 |
| | EfficientNet V2-S | 0.5553 | 0.7370 | 0.4020 | 3.3274 |
| | MobileNet V2 | 0.6340 | 0.6681 | 0.4681 | 3.1361 |
| | MobileNet V3 | 0.6770 | 0.6278 | 0.4487 | 3.1406 |
| | ResNet18 | **0.7456** | **0.5585** | 0.4995 | 3.0449 |
| | ResNet50 | 0.6703 | 0.6335 | **0.5568** | **2.8650** |
| | ShuffleNet V2 | 0.5684 | 0.7253 | 0.5414 | 2.8916 |
| Our models | VI-SABlock→CBAM | 0.7429 | 0.5568 | 0.4553 | 3.1734 |
| | VI-SABlock→ECA | 0.7324 | 0.5698 | 0.4130 | 3.2863 |
| | VI-SABlock→SE | 0.7198 | 0.5825 | 0.5377 | 2.9214 |
| | w/o VI-SABlock | 0.7316 | 0.5723 | 0.5326 | 2.9268 |
| | Unfrozen | 0.8070 | 0.4861 | 0.6520 | 2.5427 |
| | **Finetuned** | **0.8123** | **0.4796** | **0.6846** | **2.4222** |

LAI

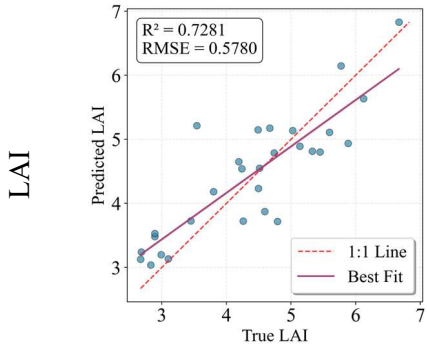 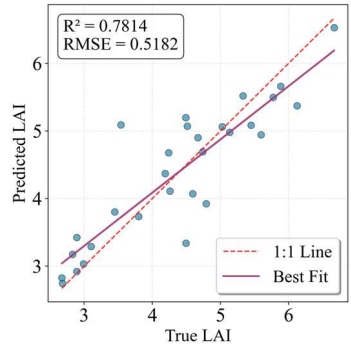 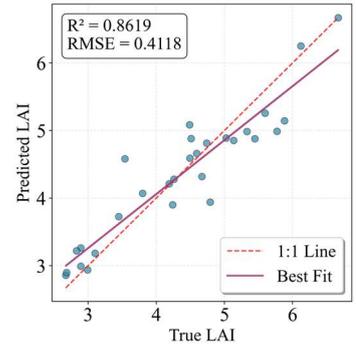

(a) SVR (VIs+TFs)      (b) ResNet 18      (c) Ours

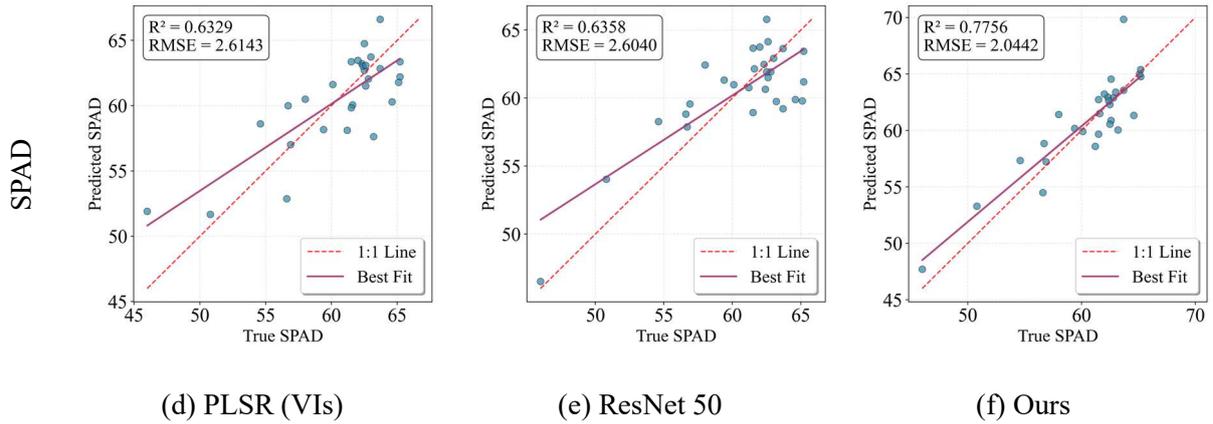

| (d) PLSR (VIs) | (e) ResNet 50 | (f) Ours |

**Fig. 6.** Predicted versus true LAI/SPAD values on the test set using the top-performing run of the best model in each category.

*4.2.1. Performance of traditional machine learning models*

In the traditional machine learning models, SVR combined with VIs and TFs achieved the best performance for LAI estimation ($R^2 = 0.7278$, RMSE = 0.5782), whereas PLSR performed the worst. Overall, incorporating TFs improved the LAI estimation accuracy across all models, with XGBoost showing the largest gain. This indicates a strong nonlinear relationship between LAI and handcrafted features, and demonstrates that TFs provide complementary information that is largely orthogonal to VIs.

In contrast, for SPAD estimation, PLSR, RFR, and XGBoost exhibited comparable performance when using only VIs, while SVR performed the poorest. After adding TFs, none of the models showed notable improvements, and the $R^2$ of PLSR and SVR significantly decreased ($p < 0.05$). This may be because, in certain growth stages, TFs exhibit weak relevance to SPAD, and indiscriminately incorporating TFs may introduce redundancy or noise (Li et al., 2020; Longfei et al., 2023). Because stage-specific modeling was not employed, the mixing of samples across different phenological phases may have further amplified these adverse effects. Although ensemble-based models such as RFR and XGBoost are more robust to redundant features, these findings highlight the inherent complexity of handcrafted feature engineering and the limitations of manually designed features in cross-stage vegetation trait estimation.

*4.2.2. Performance of traditional baseline deep learning models*

Among the baseline deep learning models, their performance on LAI estimation varies substantially. ResNet18 achieves the highest accuracy ($R^2 = 0.7456$, RMSE = 0.5585), outperforming all machine learning approaches and other deep learning baselines, with an $R^2$ significantly higher than the VIs+TFs-based SVR (p = 0.035). In contrast, EfficientNet V2-S yields the poorest results. For SPAD estimation, ResNet50 attains the best accuracy ($R^2$ = 0.5568, RMSE = 2.8650), whereas EfficientNet V2-S again performs the worst.

The ResNet series use residual connections to preserve stable gradients and extract fine-grained canopy textures, which are crucial for LAI estimation under saturation. While the deeper ResNet50 may lose some spatial detail, its higher capacity benefits SPAD prediction, which relies more on global channel-structural patterns than on local textures. In contrast, EfficientNet V2 relies heavily on compound scaling and architecture tuning derived from RGB natural-image statistics in ImageNet dataset (Tan & Le, 2021). Such design assumptions do not align with the statistical properties of VI imagery. Consequently, RGB-oriented backbone architectures, when used without task specific modifications, are suboptimal for remote sensing based LAI and SPAD estimation.

*4.2.3. Performance of our models and ablation experiment*

To evaluate the practical effectiveness of our approach, ablation experiments were conducted in which the VI-SABlock was removed from the backbone network and replaced with CBAM, ECA, or SE modules. Under fully supervised learning, the backbone network without any attention module already achieved reasonable performance in LAI estimation ($R^2$ = 0.7316, RMSE = 0.5723), with only minor improvements observed when CBAM or ECA was incorporated. VI-SABlock substantially enhanced LAI estimation accuracy ($R^2$ = 0.8070, RMSE = 0.4861), surpassing all other supervised baselines, with $R^2$ significantly higher than ResNet18 (p = 0.0002). For SPAD estimation, CBAM, ECA, and SE modules yielded no notable improvement, whereas MCVI-SANet achieved the best performance ($R^2$ = 0.6520, RMSE = 2.5427), showing a significant increase in $R^2$ over the plain backbone network (p = 0.0329).

When adopting a semi-supervised learning scheme, MCVI-SANet further improved performance in both LAI ($R^2$ = 0.8123, RMSE = 0.4796) and SPAD ($R^2$ = 0.6846, RMSE = 2.4222), outperforming all traditional machine learning and deep learning baselines. In particular, MCVI-SANet achieved an 8.95% increase in LAI estimation $R^2$ over ResNet18 and an 8.17% increase in SPAD estimation $R^2$ over PLSR. Moreover, in repeated experiments, the best results achieved by MCVI-SANet were an $R^2$ of 0.8619 and an RMSE of 0.4118 for LAI, and an $R^2$ of 0.7756 and an RMSE of 2.0442 for SPAD (Fig. 6). These optimal outcomes represent improvements of 10.3% over the best LAI baseline (ResNet18, $R^2$ = 0.7814) and 22% over the best SPAD baseline (ResNet50, $R^2$ = 0.6358). These results demonstrate the effectiveness of the proposed VI-SABlock and the VICReg-based self-supervised strategy for LAI and SPAD estimation.

*4.2.4. Deep learning model size and inference speed*

In addition to estimation accuracy, the average inference time, throughput, and parameters and model size were also measured for each deep learning model (Table 9). All models were evaluated on the CPU using 4 threads, with each single-sample forward inference repeated 100 times. Notably, ResNet50, despite having the largest number of parameters, achieves a moderate inference speed due to its simple residual design and efficient convolutional implementation. In contrast, our model has an extremely small parameter footprint (0.10 M) and model size (0.46 MB) yet maintains competitive inference time (17.05 ms). MobileNet V3 and ShuffleNet V2 exhibit the fastest inference due to their design for mobile efficiency, highlighting that parameter count alone does not linearly determine throughput on CPU.

**Table 9**
Comparison of model size and inference performance among deep learning models

| Model | Average inference time (ms) | Throughput (samples/s) | Parameters (M) | Model size (MB) |
|---|---|---|---|---|
| EfficientNet B0 | 18.12 | 55.20 | 4.05 | 15.74 |
| EfficientNet V2-S | 51.08 | 19.58 | 20.22 | 77.99 |
| MobileNet V2 | 14.16 | 70.61 | 2.27 | 8.88 |
| MobileNet V3 | 13.13 | 76.14 | 3.00 | 11.65 |
| ResNet18 | 14.15 | 70.67 | 11.22 | 42.87 |
| ResNet50 | 32.22 | 31.04 | 23.60 | 90.33 |
| ShuffleNet V2 | **10.08** | **99.22** | 0.38 | 1.57 |
| Ours | 17.05 | 58.64 | **0.10** | **0.46** |

## 5. Discussion

*5.1. Analysis of the working mechanism of VI-SABlock*

*5.1.1. Feature discrimination advantage of STD under VI saturation*

When the wheat canopy approaches closure, VIs such as NDVI typically enter a saturated range, and their mean values exhibit reduced sensitivity to changes in biomass. Experiments demonstrate that the STD retains strong discriminative power under these saturation conditions

(Fig. 7). For instance, considering NDVI, all plot images were selected from both labeled and unlabeled datasets with NDVI mean values greater than 0.8, which accounts for 80.52% of all images. Statistical analysis of the labeled subset indicates that the corresponding LAI values exceed 2.5, confirming the NDVI saturation state.

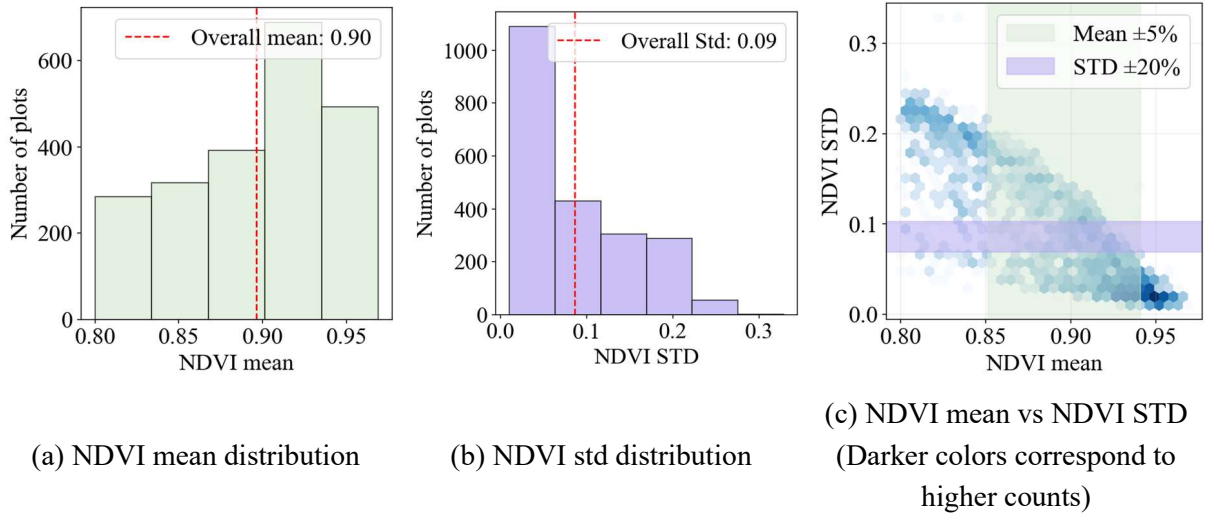

(a) NDVI mean distribution  (b) NDVI std distribution  (c) NDVI mean vs NDVI STD (Darker colors correspond to higher counts)

**Fig. 7.** Comparison of plot-level NDVI mean and STD distributions when NDVI > 0.8.

As shown in Fig. 7(a), the NDVI means of these images are highly concentrated, indicating limited discriminative capability of the mean under saturation. In contrast, the STD exhibits considerable variation (Fig. 7(c)). Quantitative results indicate that the coefficient of variation for the mean is only 4.94%, whereas that for STD reaches 76.31%, confirming that STD provides superior feature discrimination in saturated conditions. This finding is consistent with the results reported by Qiao, Zhao, et al. (2022), who demonstrated that when TFs are incorporated, the STD of VIs within the selected region exhibits the strongest correlation with LAI.

The underlying physical mechanism is that STD captures spatial heterogeneity, directly reflecting canopy-level interactions such as light-shadow patterns, leaf angle distributions, and micro-stress patches. Even when mean values are rendered ineffective by saturation, STD continues to convey fine-grained canopy structural differences, providing critical complementary information, which is consistent with the findings of Le Maire et al. (2006). This property is not captured by modules relying solely on GAP or MAP, such as ECA, CBAM or SE, highlighting the unique advantage of VI-SABlock in addressing VI saturation.

*5.1.2. Visualization of attention mechanisms in VI-SABlock*

To further elucidate the response mechanism of VI-SABlock to multi-channel VI inputs, this study implemented attention visualization based on Grad-CAM. This approach helps to prevent the bias in attention weights introduced by BN and feature expansion layers. Specifically, the convolutional expansion layer at the end of the module was used to generate heatmaps, which characterize the contribution of different spatial regions to the prediction. To eliminate bias caused by scale inconsistencies among VIs, each input channel was first normalized. The normalized channels were then multiplied pixel-wise by the heatmap. After that, the results were aggregated along the spatial dimension to derive the channel attention weights, therefore the sum of all channel weights is 1. The resulting visualizations effectively reveal the module's differential focus on VI channels, thereby enhancing interpretability.

Experiments were conducted on wheat plots with varying LAI and SPAD values (Fig. 8), where the SPAD content decreases progressively from plots Fig. 8(a) to Fig. 8(e). Spatially, the module consistently concentrates attention on vegetated areas (warmer colors indicate higher attention) with minimal interference from bare soil backgrounds. Notably, in Fig. 8(a), the right portion of the plot is heavily shadowed and visually indistinguishable in the RGB image; however, VI-SABlock still accurately highlights vegetation-related patterns by leveraging VI information. In addition, in Fig. 8(e), the attention is mainly focused on the green areas of the field rather than the yellowing areas. These results demonstrate that VI-SABlock can effectively integrate VI and spatial attention representations, exhibiting remarkable effectiveness under complex illumination and canopy structure conditions.

In the channel dimension, attention weights exhibit clear adaptive adjustment trends with decreasing SPAD. Overall, the module tends to assign higher weights to saturation-resistant indices such as GNDVI and NDRE. NDVI still receives relatively high weights in some plots, likely due to its sensitivity to high-frequency textures in the upper canopy. With decreasing SPAD, certain indices display a consistent variation pattern. For example, the increasing weight of MCARI suggests that VI-SABlock enhances reliance on chlorophyll-sensitive VIs to maintain accurate SPAD perception. Additionally, when a complex soil background is present (e.g., Fig. 8(c)), VIs such as OSAVI and SAVI show elevated weights, indicating strong robustness against background interference. Conversely, VIs with large dynamic ranges, such as RVI and CIgreen, receive lower attention weights to prevent dominance effects that could suppress the contributions of other indices.

In summary, these findings confirm that VI-SABlock dynamically selects informative VIs according to canopy spectral characteristics. Its channel and spatial attention mechanisms then work together to enhance feature representation. This synergy leads to significantly improved performance under dense canopy and complex background conditions.

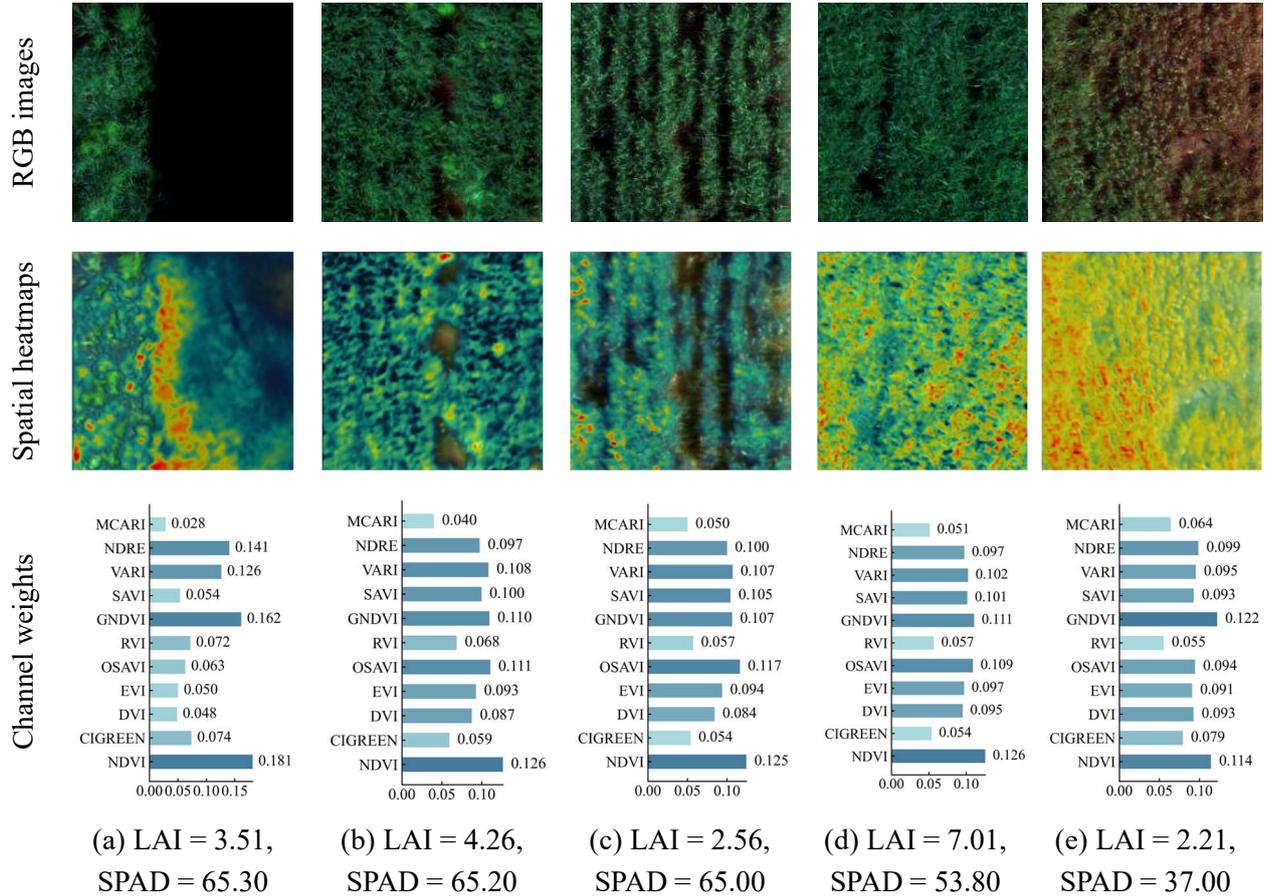

**Fig. 8.** RGB images, spatial heatmaps, and channel weights under different plots. The spatial heatmap is generated by overlaying heatmap colors on the original RGB imagery.

### 5.2. Applicability of VICReg for LAI and SPAD estimation

In this study, VICReg-based self-supervised features effectively capture both the spatial structure and the multi-channel VI information of the winter wheat canopy. It maintains embedding stability and prevents feature collapse, which makes it more robust when samples are highly similar or share complex visual patterns (Bardes et al., 2022). Specifically, the invariance regularization ensures robustness to canopy image augmentations, enabling the model to extract continuous representations. The variance regularization prevents representation collapse, ensuring sufficient separation among samples. The covariance regularization promotes feature diversity and complementarity. Moreover, the lightweight model can quickly learn effective representations from a small set of self-supervised samples. These properties explain the observed stability and high accuracy of MCVI-SANet across multiple growth stages and field plots.

MAE-based pretraining was also explored, but it was found to require high-parameter models and large training datasets to achieve fine-grained reconstruction, as it is built upon the vision Transformer (ViT) architecture (He et al., 2022). MAE consumed excessive computational resources, and training was extremely slow, with suboptimal reconstruction quality on the test data (Fig. 9). Furthermore, using the frozen ViT encoder outputs as features for a fully connected network to predict LAI or SPAD failed to converge. Likewise, contrastive learning methods relying on negative samples were unsuitable due to high sample similarity and continuous target values.

In contrast, VICReg-based MCVI-SANet, equipped with a simple two-layer fully connected network is sufficient for rapid convergence while achieving high-accuracy predictions. The embeddings generated by the expander of the self-supervised MCVI-SANet across training epochs were also analyzed (Fig. 10). The results show that no embedding dimensions exhibited collapse (defined as variance below $1\times10^{-3}$) even at the first epoch, and the variance distribution converged to a stable state by around epoch 200. Moreover, the embedding variances at epoch 196 (Fig. 11) indicate that almost all dimensions contain informative and highly discriminative representations. These results demonstrates the strong suitability and stability of VICReg for representation learning on multi-channel VI imagery.

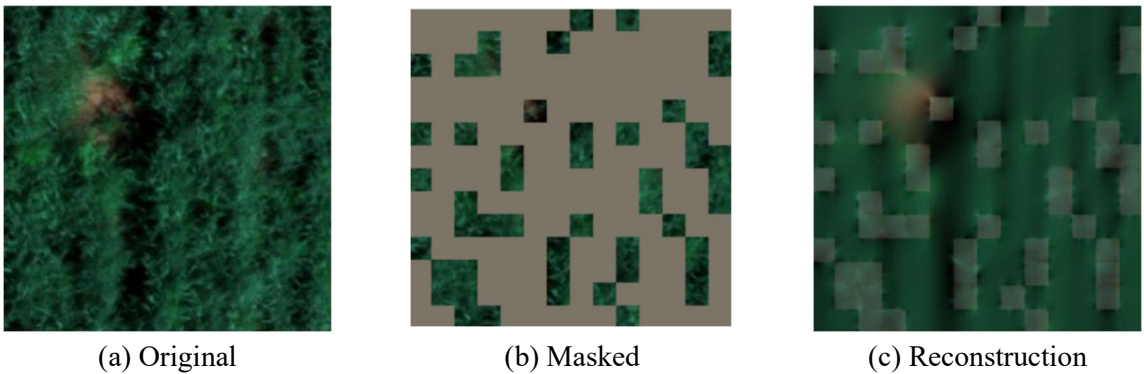

(a) Original　　　　　　　　(b) Masked　　　　　　　(c) Reconstruction

**Fig. 9.** Reconstruction results of winter wheat canopy RGB images using MAE, with a 75% masking ratio as recommended by MAE.

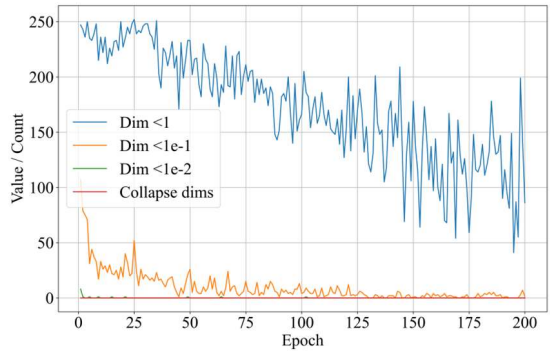 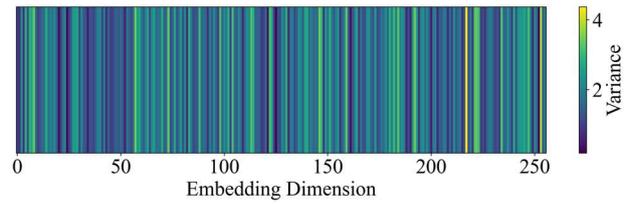

**Fig. 10.** Trend of embedding dimension variances.

**Fig. 11.** Visualization of embedding dimension variances at epoch 196. Brighter colors indicate higher variance.

*5.3. Estimation analysis for all plots*

The LAI and SPAD values of all wheat plots across growth stages were estimated based on MCVI-SANet, and the spatial-temporal heatmaps of predictions, ground truth, and estimation errors were generated (Fig. 12). The results show that the high LAI areas were mainly located in the southern-central region of the wheat field, which was consistent between the predicted and measured heatmaps. The LAI error distribution further indicates that five consecutive plots exhibited high errors during the flowering stage. Field inspection confirmed structural disturbance and emerging weeds in this area, which resulted in overestimated LAI measurements, while MCVI-SANet provided lower estimates. Since all five plots belonged to the training set, the impact on validation and test performance was minimal. Nevertheless, the model's ability to maintain reasonable estimates despite noisy training data demonstrates MCVI-SANet's robustness to data noise.

For SPAD, the overall spatial difference was less pronounced, but low-value areas emerged at field boundaries during the grain-filling period, accompanied by increased prediction errors. In addition, the average SPAD value during the booting stage was lower than those during the neighboring jointing and heading stages, consistent with the patterns observed in NDVI, NDRE, RVI, and several other VIs (Fig. 13). This phenomenon may be partly attributed to nutrient translocation from the leaves to the developing spikes and the senescence of older leaves, while the flag leaves had not fully expanded. In addition, the booting stage may have experienced water and nitrogen stress, as precipitation was infrequent and the average temperature was relatively high during this period.

Moreover, after the flowering stage, the wheat canopy exhibited pronounced leaf yellowing and curling heterogeneity. Since the SPAD-502 measures only a limited number of sampled leaves per plot, whereas LAI measurements cover nearly the entire canopy, the measurement uncertainty of SPAD increased, resulting in larger prediction errors during this period.

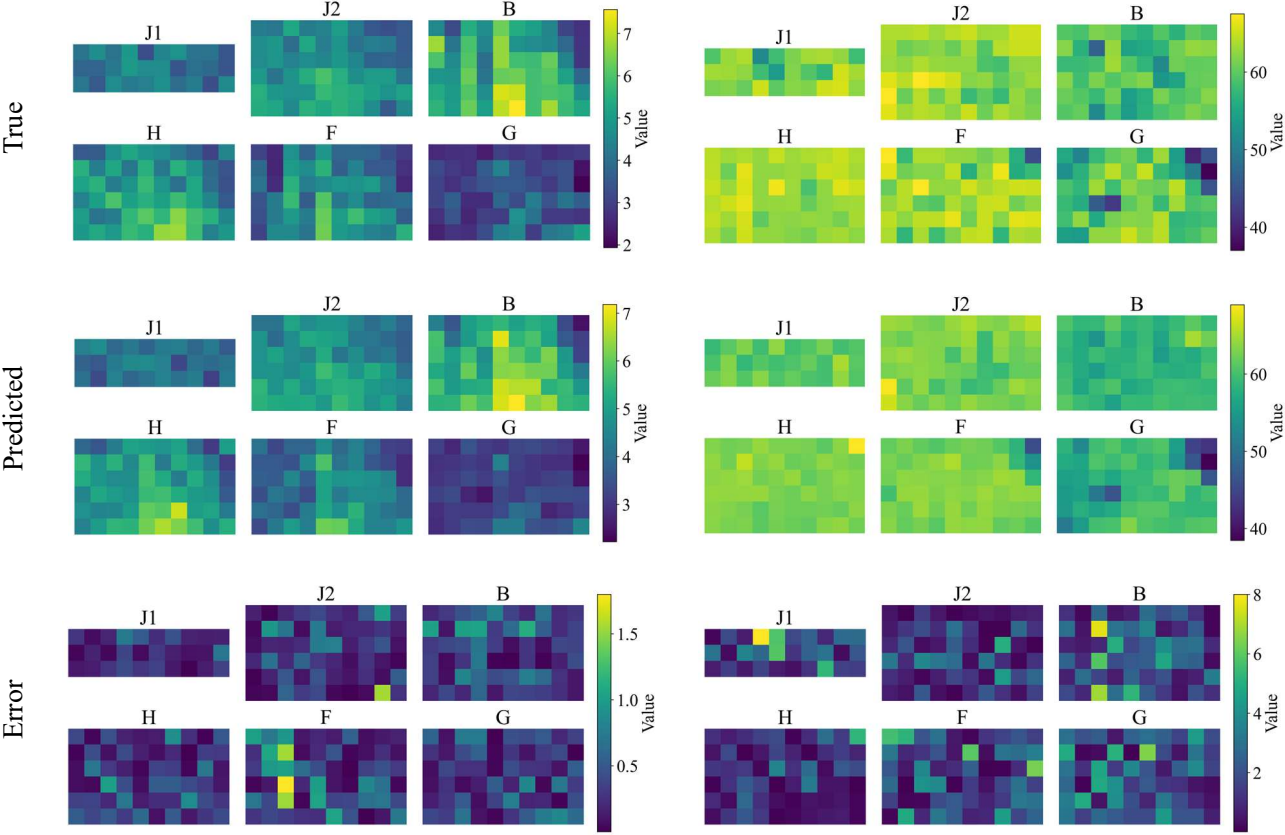

(a) LAI distribution　　　　　　　　　　　(b) SPAD distribution

**Fig. 12.** Field distribution of true value, predicted value, and error value for LAI and SPAD across all growth stages.

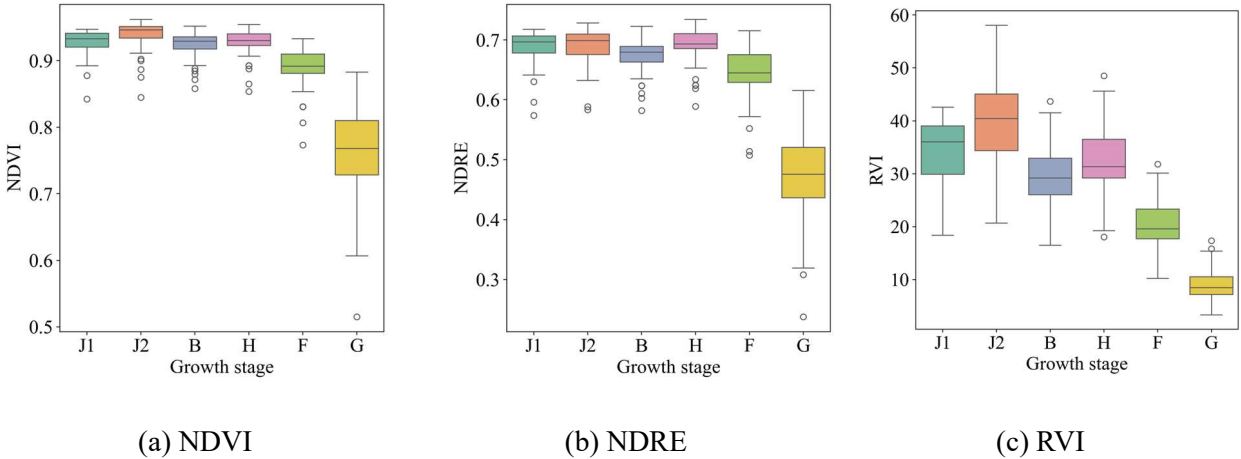

(a) NDVI　　　　　　　　(b) NDRE　　　　　　　　(c) RVI

**Fig. 13.** Boxplots of NDVI, NDRE, and RVI at different growth stages

*5.4. Limitations of MCVI-SANet and future perspectives*

Despite the superior accuracy and efficiency of MCVI-SANet in estimating LAI and SPAD, several limitations remain. First, the model's response to extreme biotic and abiotic stresses has not been evaluated, nor has it been assessed for other crops. Future studies could collect data for different crops under various stress conditions to develop more generalizable models. Second, relies solely on multi-channel VI imagery may not capture the full complexity of genotype-environment interactions. Multi-source fusion frameworks, integrating thermal infrared and SAR data, as well as temporal models such as LSTM, could enable dynamic growth prediction across the phenological cycle. Finally, although MCVI-SANet is lightweight, its inference speed still lags behind traditional machine learning methods. Future work may adopt model compression techniques, neural architecture search, or knowledge distillation for real-time deployment. Additionally, the limited self-supervised training set constrains performance. Expanding the dataset and incorporating multi-crop data could fully exploit SSL's potential.

**6. Conclusion**

This study introduces MCVI-SANet, a lightweight semi-supervised vision model. It is designed to address VI saturation and limited labeled samples in estimating LAI and SPAD during the dense canopy stage of winter wheat. The model integrates the VI-SABlock for adaptive spectral-spatial feature enhancement, and employs a VICReg-based semi-supervised method to effectively leverage unlabeled data. This approach improves generalization while maintaining efficiency with only 0.10M parameters. Experimental results confirm its state-of-the-art performance. MCVI-SANet achieves the highest accuracy in both LAI and SPAD estimation, with an 8.95% increase in the average LAI estimation $R^2$ and an 8.17% increase in the average SPAD estimation $R^2$ over the best-performing baselines. The findings highlight the potential of combining physical priors with deep learning to overcome challenges in agricultural remote sensing. Although developed for winter wheat, the VICReg-based semi-supervised strategy and network architecture are generally applicable to other crops with similar multispectral data. Future research could explore its application across diverse crops and environmental conditions to further advance precision agriculture.

**CRediT authorship contribution statement**

**Zhiheng Zhang:** Writing – original draft, Conceptualization, Methodology, Software, Investigation, Visualization. **Jiajun Yang:** Investigation. **Hong Sun:** Writing – review & editing. **Dong Wang:** Writing – review & editing. **Honghua Jiang:** Writing – review & editing, Data curation. **Yaru Chen:** Writing – review & editing. **Tangyuan Ning:** Writing – review & editing, Resources.

**Declaration of Competing Interest**

The authors declare that they have no known competing financial interests or personal relationships that could have appeared to influence the work reported in the paper.

**Acknowledgements**

This work was supported by the R&D and Integrated Demonstration of Key Technologies for Precise Control of Annual Water and Fertilisation in Wheat and Maize (2019JZZY010716), the National Key R&D Program Strategic Science and Technology Innovation Cooperation key special project "Cooperative Research on AI-Enhanced Soil and Crop Sensing Technology" (2025YFE0209000), and the National Key R&D Program of China (2023YFD2001400).

**Data availability**

The data used in this paper will be made available on request.